\documentclass[10pt,twocolumn,letterpaper]{article}

\usepackage{cvpr}
\usepackage{times}
\usepackage{epsfig}
\usepackage{graphicx}
\usepackage{amsmath}
\usepackage{amssymb}

\usepackage{siunitx}
\usepackage{threeparttable}

\usepackage[pagebackref=true,breaklinks=true,letterpaper=true,colorlinks,bookmarks=false]{hyperref}

\cvprfinalcopy %

\def\cvprPaperID{1445} %

\begin{document}

\title{HigherHRNet: Scale-Aware Representation Learning for\\Bottom-Up Human Pose Estimation}

\author{Bowen Cheng$^{1}$, Bin Xiao$^{2}$, Jingdong Wang$^{2}$, Honghui Shi$^{1,3}$, Thomas S. Huang$^{1}$, Lei Zhang$^{2}$\\
\\
{$^1$UIUC, $^2$Microsoft, $^3$University of Oregon}}

\maketitle

\begin{abstract}
Bottom-up human pose estimation methods have difficulties in predicting the correct pose for small persons due to challenges in scale variation.
In this paper, 
we present \textbf{HigherHRNet}: a novel bottom-up human pose estimation method for learning scale-aware representations using high-resolution feature pyramids.
Equipped with multi-resolution supervision for training
and multi-resolution aggregation  for inference,
the proposed approach is able to solve
the scale variation challenge 
in \emph{bottom-up multi-person} pose estimation
and localize keypoints more precisely, especially for small person.
The feature pyramid in HigherHRNet
consists of feature map 
outputs from HRNet 
and upsampled higher-resolution outputs
through a transposed convolution.
HigherHRNet outperforms the previous best bottom-up method by $2.5\%$ AP for medium person on COCO test-dev, showing its effectiveness in handling scale variation. Furthermore, HigherHRNet achieves new state-of-the-art result on COCO test-dev ($70.5\%$ AP) without using refinement or other post-processing techniques, surpassing all existing bottom-up methods. HigherHRNet even surpasses all top-down methods on CrowdPose test ($67.6\%$ AP), suggesting its robustness in crowded scene. 
The code and models are available at~\url{https://github.com/HRNet/Higher-HRNet-Human-Pose-Estimation}.
\end{abstract}

\section{Introduction}
2D human pose estimation aims at localizing human anatomical keypoints (\emph{e.g.}, elbow, wrist, \emph{etc.}) or parts. As a fundamental technique to human behavior understanding, it has received increasing attention in recent years.

\begin{figure}
    \centering
    \begin{minipage}[b]{0.99\linewidth}
    \centering
    \vspace{2mm}
    \includegraphics[width=0.9\linewidth]{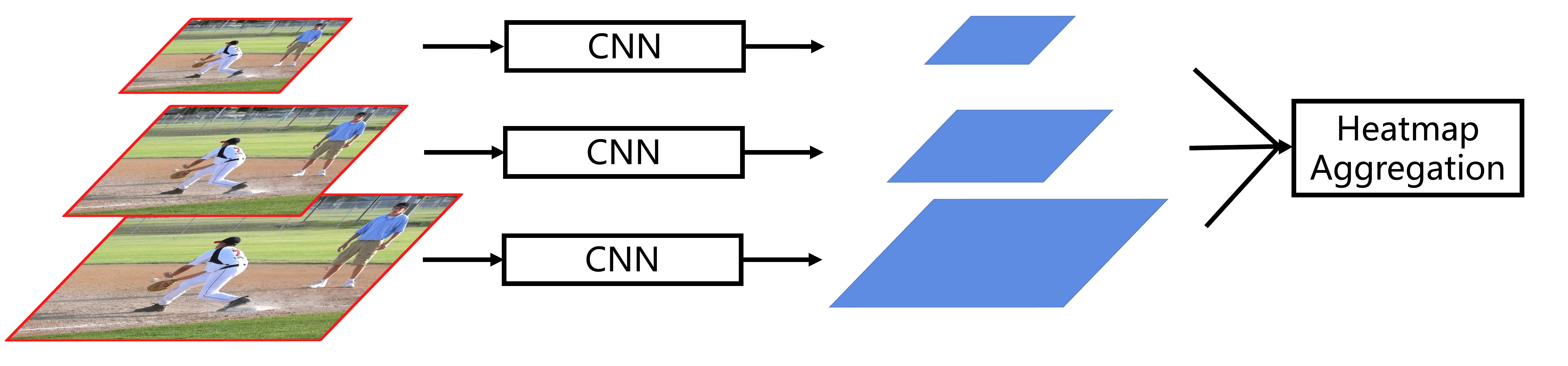}
    \vspace{2mm}\\
    {\footnotesize(a) Image pyramid.}
    \vspace{2mm}\\
    \end{minipage}
    \begin{minipage}[b]{0.99\linewidth}
    \centering
    \includegraphics[width=0.9\linewidth]{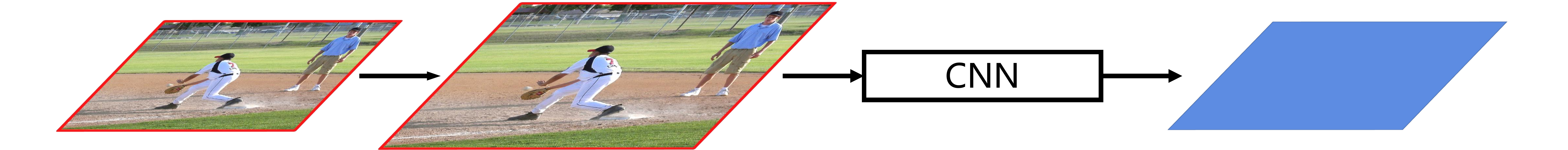}
    \vspace{2mm}\\
    {\footnotesize(b) Upsampling input.}
    \end{minipage}
    \begin{minipage}[b]{0.99\linewidth}
    \centering
    \vspace{2mm}
    \includegraphics[width=0.9\linewidth]{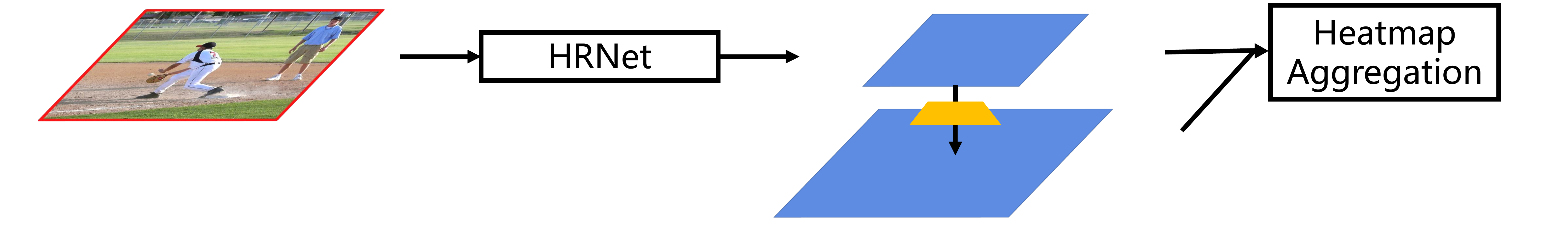}
    \vspace{2mm}\\
    {\footnotesize(c) Our approach.}
    \end{minipage}
    \vspace{2mm}
    \caption{(a) Using image pyramid for heatmap prediction~\cite{papandreou2018personlab,newwell2017associative}. (a) Generating higher resolution and  spatially more accurate heatmaps by upsampling image. Recent work PersonLab~\cite{papandreou2018personlab} relies on enlarging input image size to generate high quality feature maps.  (c) Our HigherHRNet uses high resolution feature pyramid.
    }
    \vspace{-4mm}
    \label{fig:intro}
\end{figure}

Current human pose estimation methods can be categorized into \emph{top-down} methods and \emph{bottom-up} methods. Top-down methods \cite{papandreou2017towards,chen2018cascaded,he2017mask,xiao2018simple,sun2019deep,WangSCJDZLMTWLX19,sun2018integral,he2017mask} take a dependency on person detector to detect person instances each with a bounding box and then reduce the problem to a simpler task of single person pose estimation. As top-down methods can normalize all the persons to approximately the same scale by cropping and resizing the detected person bounding boxes, they are generally less sensitive to the scale variance of persons. Thus, state-of-the-art performances on various multi-person human pose estimation benchmarks are mostly achieved by top-down methods. However, as such methods rely on a separate person detector and need to estimate pose for every person individually, they are normally computationally intensive and not truly end-to-end systems. By contrast, bottom-up methods \cite{cao2017realtime,newwell2017associative,papandreou2018personlab,kreiss2019pifpaf} start by localizing identity-free keypoints for all the persons in an input image through predicting heatmaps of different anatomical keypoints, followed by grouping them into person instances. This strategy effectively makes bottom-up methods faster and more capable of achieving real-time pose estimation. However, because bottom-up methods need to deal with scale variation, there still exists a large gap between the performances of bottom-up and top-down methods, especially for small scale persons.

There are mainly two challenges in predicting keypoints of small persons. One is dealing with scale variation, \emph{i.e.} to improve the performance of small person without sacrificing the performance of large persons. The other is generating a high-resolution heatmap with high quality for precise localizing keypoints of small persons. 
Previous bottom-up methods \cite{cao2017realtime,newwell2017associative,papandreou2018personlab,kreiss2019pifpaf} mainly focus on grouping keypoints and simply use a single resolution of feature map that is 1/4 of the input image resolution to predict the heatmap of keypoints. These methods neglect the challenge of scale variation and rely on image pyramid during inference (Figure~\ref{fig:intro} (a)). Feature pyramids are basic components for handling scale variation, however, smaller resolution feature maps in a top-down feature pyramid usually suffer from the second challenge. PersonLab~\cite{papandreou2018personlab} generates high-resolution heatmaps by increasing the input resolution (Figure~\ref{fig:intro} (b)). Although the performance of small persons increases consistently as input resolution, the performance of large persons begin decreasing when input resolution is too large. To solve these challenges, it is crucial to generate spatially more accurate and scale-aware heatmaps for bottom-up keypoint prediction in a natural and simple way without sacrificing computational cost.

In this paper, we propose a Scale-Aware High-Resolution Network (HigherHRNet) to address these challenges. HigherHRNet generates high-resolution heatmaps by a new high-resolution feature pyramid module. Unlike the traditional feature pyramid that starts from 1/32 resolution and uses bilinear upsampling with lateral connection to gradually increases feature map resolution to 1/4, high-resolution feature pyramid directly starts from 1/4 resolution which is the highest resolution feature in the backbone and generates even higher-resolution feature maps with deconvolution (Figure~\ref{fig:intro} (c)). We build the high-resolution feature pyramid on the 1/4 resolution path of HRNet~\cite{sun2019deep,WangSCJDZLMTWLX19}, to make it efficient. To make HigherHRNet capable of handling scale variation, we further propose a Multi-Resolution Supervision strategy to assign training target of different resolutions to the corresponding feature pyramid level. Finally, we introduce a simple Multi-Resolution Heatmap Aggregation strategy
during inference to generate scale-aware high-resolution heatmaps.

We validate our method on the challenging COCO keypoint detection dataset \cite{lin2014microsoft} and demonstrate superior keypoint detection performance. Specifically, HigherHRNet achieves AP of $70.5\%$ on COCO2017 test-dev \emph{without any post processing}, outperforming all existing bottom-up methods by a large margin. Furthermore, we observe that most of the gain comes from medium person (there is no small person annotation for the keypoint detection task), HigherHRNet outperforms the previous best bottom-up method by $2.5\%$ AP for medium persons without sacraficing the performance of large persons ($+0.3\%$ AP). This observation verifies HigherHRNet is indeed solving the scale variation challenge. We also provide a solid baseline for bottom-up methods on the new CrowdPose~\cite{li2019crowdpose} dataset. Our HigherHRNet achieves AP of $67.6\%$ on CrowdPose test, surpassing all existing methods. This result suggests bottom-up methods naturally have the advantages in the crowded scene.

To summarize our contributions:
\begin{itemize}
\setlength\itemsep{0em}
\item
We attempt to address the scale variation challenge,
which is rarely studied before in bottom-up multi-person pose estimation.
\item
We propose a HigherHRNet that generates high-resolution feature pyramid with multi-resolution supervision in the training stage and multi-resolution heatmap aggregation in the inference stage to predict scale-aware high-resolution heatmaps that are beneficial for small persons.
\item
We demonstrate the effectiveness of our HigherHRNet on the challenging COCO dataset. Our model outperforms all other bottom-up methods. We especially observe a large gain for medium persons.
\item
We achieve a new state-of-the-art result on the CrowdPose dataset, suggesting bottom-up methods are more robust to the crowded scene over top-down methods.
\end{itemize}

\section{Related works}
\paragraph{Top-down methods.} Top-down methods \cite{xiao2018simple,sun2019deep,WangSCJDZLMTWLX19,papandreou2017towards,he2017mask,huang2017coarse,fang2017rmpe,chen2018cascaded,newell2016stacked} detect the keypoints of a single person within a person bounding box. The person bounding boxes are usually generated by an object detector \cite{ren2015faster,lin2017feature,cheng2018revisiting,cheng2018decoupled}. Mask R-CNN \cite{he2017mask} directly adds a keypoint detection branch on Faster R-CNN \cite{ren2015faster} and reuses features after ROIPooling. G-RMI~\cite{papandreou2017towards} and the following methods further break top-down methods into two steps and use separate models for person detection and pose estimation.
\vspace{-2mm}
\paragraph{Bottom-up methods.} Bottom-up methods \cite{pishchulin2016deepcut,insafutdinov2016deepercut,iqbal2016multi,cao2017realtime,newwell2017associative} detect identity-free body joints for all the persons in an image and then group them into individuals. OpenPose~\
\cite{cao2017realtime} uses a two-branch multi-stage netork with one branch for heatmap prediction and one branch for grouping. OpenPose uses a grouping method named part affinity field which learns a 2D vector field linking two keypoints. Grouping is done by calculating line integral between two keypoints and group the pair with the largest integral. Newell \emph{et al.} \cite{newwell2017associative} use stacked hourglass network \cite{newell2016stacked} for both heatmap prediction and grouping. Grouping is done by a method named associate embedding, which assigns each keypoint with a ``tag'' (a vector representation) and groups keypoints based on the $\textit{l}_{2}$ distance between tag vectors. PersonLab~\cite{papandreou2018personlab} uses dilated ResNet~\cite{he2016deep} and groups keypoints by directly learning a 2D offset field for each pair of keypoints. PifPaf~\cite{kreiss2019pifpaf} uses a Part Intensity Field (PIF) to localize body parts and a Part Association Field (PAF) to associate body parts with each other to form full human poses.

\begin{figure*}[t]
    \centering
    \includegraphics[width=0.98\textwidth]{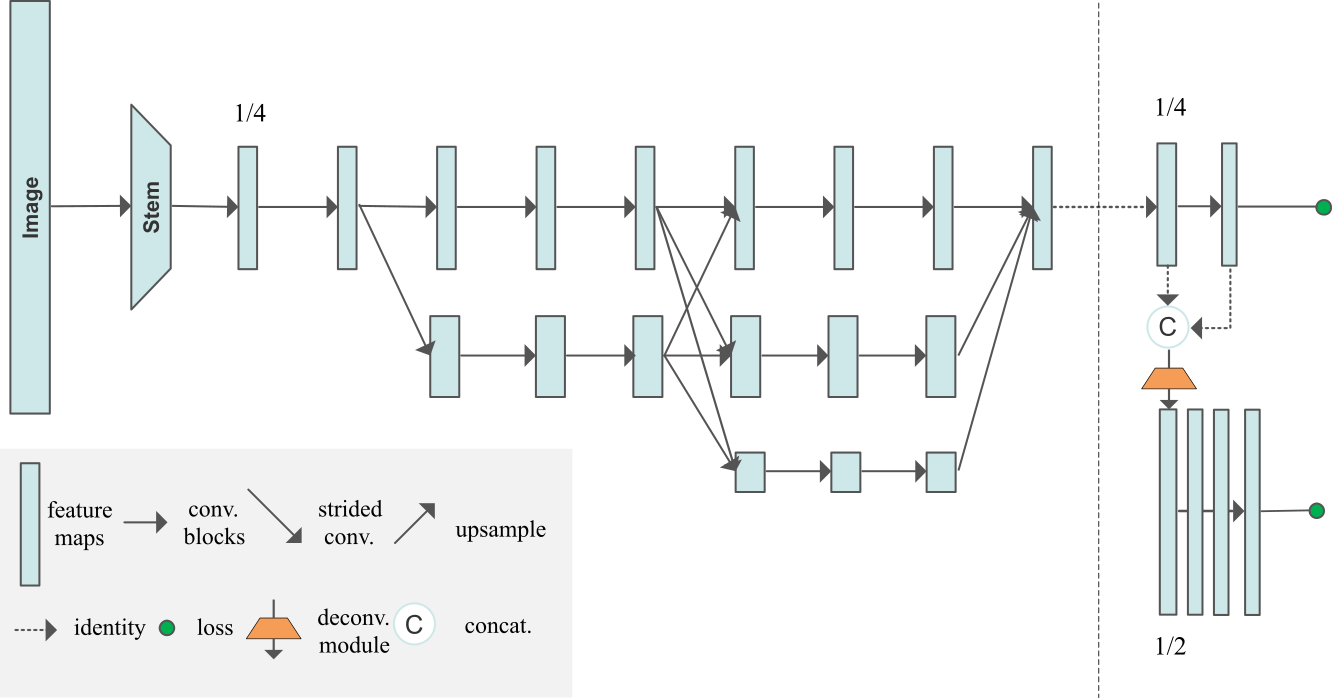}
    \caption{An illustration of HigherHRNet. The network uses HRNet~\cite{sun2019deep,WangSCJDZLMTWLX19} as backbone, followed by one or more deconvolution modules to generate multi-resolution and high-resolution heatmaps. Multi-resolution supervision is used for training. More details are given in Section~\ref{sec:approach}.}
    \vspace{-4mm}
    \label{fig:hrnet}
\end{figure*}
\vspace{-4mm}
\paragraph{Feature pyramid.} Pyramidal representation has been widely adopted in recent object detection and segmentation frameworks to handle scale variation. SSD~\cite{liu2016ssd} and MS-CNN~\cite{cai2016unified} predict objects at multiple layers of the network without merging features. Feature pyramid networks~\cite{lin2017feature} extend the backbone model with a top-down pathway that gradually recovers feature resolution from 1/32 to 1/4, using bilinear upsampling and lateral connection. The motivation in common is to let features from different pyramid level to predict instances of different scales. However, this pyramidal representation is less explored in bottom-up multi-person pose estimation. In this work, we design a high-resolution feature pyramid that extend the pyramid to a different direction, starting from 1/4 resolution feature and generate pyramid of features with higher resolution.
\vspace{-4mm}
\paragraph{High resolution feature maps.} There are mainly 4 methods to generate high resolution feature maps. (1) Encoder-decoder \cite{newell2016stacked,he2017mask,chen2018cascaded,ronneberger2015u,badrinarayanan2017segnet,lin2017refinenet,wojna2017devil,cheng2019spgnet} captures the context information in the encoder path and recover high resolution features in the decoder path. The decoder usually contains a sequence of bilinear upsample operations with skip connections from encoder features with the same resolution. (2) Dilated convolution \cite{yu2015multi,deeplabv12015,chen2018deeplabv2,chen2017deeplabv3,deeplabv3plus2018,dpc2018,liu2019auto,yang2019deeperlab,cheng2019panoptic,cheng2020panoptic} (\emph{a.k.a.} ``atrous'' convolution) is used to remove several stride convolutions/max poolings to preserve feature map resolution. Dilated convolution prevents losing spatial information but introduces more computational cost. (3) Deconvolution (transposed convolution) \cite{xiao2018simple} is used in sequence at the end of a network to efficiently increase feature map resolution. SimpleBaseline~\cite{xiao2018simple} demonstrates that deconvolution can generate high quality feature maps for heatmap prediction. (4) Recently, a High-Resolution Network (HRNet)~\cite{sun2019deep, WangSCJDZLMTWLX19} is proposed as an efficient way to keep a high resolution pass throughout the network. HRNet~\cite{sun2019deep,WangSCJDZLMTWLX19} consists of multiple branches with different resolutions. Lower resolution branches capture contextual information and higher resolution branches preserve spatial information. With multi-scale fusions between branches, HRNet~\cite{sun2019deep,WangSCJDZLMTWLX19} can generate high resolution feature maps with rich semantic.

We adopt HRNet~\cite{sun2019deep,WangSCJDZLMTWLX19} as our base network to generate high-quality feature maps. And we add a deconvolution module to generate higher resolution feature maps to predict heatmaps.
The resulting model is named ``Scale-Aware High-Resolution Network'' (HigherHRNet). As both HRNet~\cite{sun2019deep,WangSCJDZLMTWLX19,WangSCJDZLMTWLX19} and deconvolution are efficient, HigherHRNet is an efficient model for generating higher resolution feature maps for heatmap prediction.

\section{Higher-Resolution Network}
\label{sec:approach}

In this section, we introduce our proposed Scale-Aware High-Resolution Representation Learning using the HigherHRNet. Figure~\ref{fig:hrnet} illustrates the overall architecture of our method. We will firstly give a brief overview on the proposed HigherHRNet and then describe its components in details.

\subsection{HigherHRNet}

\paragraph{HRNet.} HigherHRNet uses HRNet~\cite{sun2019deep,WangSCJDZLMTWLX19} (shown in Figure~\ref{fig:hrnet}) as backbone. HRNet~\cite{sun2019deep,WangSCJDZLMTWLX19} starts with a high-resolution branch in the first stage. In every following stage, a new branch is added to current branches in parallel with $\frac{1}{2}$ of the lowest resolution in current branches. As the network has more stages, it will have more parallel branches with different resolutions and resolutions from previous stages are all preserved in later stages. An example network structure, containing 3 parallel branches, is illustrated in Figure~\ref{fig:hrnet}.

We instantiate the backbone using a similar manner as HRNet~\cite{sun2019deep,WangSCJDZLMTWLX19}. The network starts from a stem
that consists of two strided $3 \times 3$ convolutions
decreasing the resolution to $1/4$.
The $1$st stage contains $4$ residual units 
where each unit
is formed by a bottleneck with width (number of channels) $64$,
followed by one $3\times3$ convolution
reducing the width of feature maps
to $C$.
The $2$nd, $3$rd, $4$th stages
contain $1$, $4$, and $3$ multi-resolution blocks, respectively.
The widths of the convolutions
of the four resolutions 
are $C$, $2C$, $4C$, and $8C$, respectively.
Each branch in the multi-resolution group convolution
has $4$ residual units 
and each unit has two $3\times3$ convolutions in each resolution. We experiment with two networks with different capacity by setting $C$ to 32 and 48 respectively.

HRNet~\cite{sun2019deep,WangSCJDZLMTWLX19} was originally designed for top-down pose estimation. In this work, we adopt HRNet~\cite{sun2019deep,WangSCJDZLMTWLX19} to a bottom-up method by adding a $1\times1$ convolution to predict heatmaps and tagmaps similar to \cite{newwell2017associative}. We only use the highest resolution ($\frac{1}{4}$ of the input image) feature maps for prediction. Following \cite{newwell2017associative}, we use a scalar tag for each keypoint.
\vspace{-4mm}
\paragraph{HigherHRNet.} 
Resolution of the heatmap is important for predicting keypoints for small persons. Most existing human pose estimation methods predict Gaussian-smoothed heatmaps by preparing the ground truth headmaps with an unnormalized Gaussian kernel applyed to each keypoint location. Adding this Gaussian kernel helps training networks as CNNs tend to output spatially smooth responses as a nature of convolution operations. However, applying a Gaussian kernel also introduces confusion in precise localization of keypoints, especially for keypoints belonging to small persons. A trivial solution to reduce this confusion is to reduce the standard deviation of the Gaussian kernel. However, we empirically find that it makes optimization harder and leads to even worse results. 

Instead of reducing standard deviation, we solve this problem by predicting heatmaps at higher resolution with standard deviation unchanged at different resolutions. Bottom-up methods usually predict heatmaps at resolution $\frac{1}{4}$ of the input image. Yet we find this resolution is not high enough for predicting accurate heatmaps. Inspired by \cite{xiao2018simple}, which shows that deconvolution can be used to effectively generate high quality and high resolution feature maps, we build HigherHRNet on top of the highest resolution feature maps in HRNet as shown in Figure~\ref{fig:hrnet} by adding a deconvolution module as discussed in Section~\ref{sec:deconv_modules}.

The deconvolution module takes as input both features and predicted heatmaps from HRNet and generates new feature maps that are $2$ times larger in resolution than the input feature maps. A feature pyramid with two resolutions is thus generated by the deconvolution module together with the feature maps from HRNet. The deconvolution module also predicts heatmaps by adding an extra $1\times1$ convolution. We follow Section~\ref{sec:mr_supervision} to train heatmap predictors at different resolutions and use a heatmap aggregation strategy as described in (Section~\ref{sec:mr_inference}) for inference.

More deconvolution modules can be added if larger resolution is desired. We find the number of deconvolution modules is dependent on the distribution of person scales of the dataset. Generally speaking, a dataset containing smaller persons requires larger resolution feature maps for prediction and vice versa. In experiments, we find adding a single deconvolution module achieves the best performance on the COCO dataset.

\subsection{Grouping.} Recent works~\cite{newwell2017associative,law2018cornernet} have shown that grouping can be solved with high accuracy by a simple method using associative embedding \cite{newwell2017associative}. As an evidence, experimental results in~\cite{newwell2017associative}  show that using the ground truth detections with the predicted tags improves AP from $59.2$ to $94.0$ on a held-out set of 500 training images of the COCO keypoint detection dataset~\cite{lin2014microsoft}. We follow \cite{newwell2017associative} to use associative embedding for keypoint grouping. The grouping process clusters identity-free keypoints into individuals by grouping keypoints whose tags have small $l_2$ distance.

\subsection{Deconvolution Module}
\label{sec:deconv_modules}
We propose a simple deconvolution module for generating high quality  feature maps whose resolution is two times higher than the input feature maps. Following \cite{xiao2018simple}, we use a $4\times4$ deconvolution (\emph{a.k.a.} transposed convolution) followed by BatchNorm and ReLU to learn to upsample the input feature maps. Optionally, we could further add several Basic Residual Blocks \cite{he2016deep} after deconvolution to refine the upsampled feature maps. We add 4 Residual Blocks in HigherHRNet.

Different from \cite{xiao2018simple}, the input to our deconvolution module is the concatenation of the feature maps and the predicted heatmaps from either HRNet or previous deconvolution modules. And the output feature maps of each deconvolution module are also used to predict heatmaps in a multi-scale fashion.

\subsection{Multi-Resolution Supervision}
\label{sec:mr_supervision}
Unlike other bottom-up methods \cite{newwell2017associative,papandreou2018personlab,cao2017realtime} that only apply supervision to the largest resolution heatmaps, we introduce a multi-resolution supervision during training to handle scale variation. We transform ground truth keypoint locations to locations on the heatmaps of all resolutions to generate ground truth heatmaps with different resolutions. Then we apply a Gaussian kernel \emph{with the same standard deviation} (we use $\text{standard deviation}=2$ by default) to all these ground truth heatmaps. We find it important not to scale standard deviation of the Gaussian kernel. This is because different resolution of feature pyramid is suitable to predict keypoints of different scales. On higher-resolution feature maps, a relatively small standard deviation (compared to the resolution of the feature map) is desired to more precisely localize keypoints of small persons. 

At each prediction scale in HigherHRNet, we calculate the mean squared error between the predicted heatmaps of that scale and its associated ground truth heatmaps. The final loss for heatmaps is the sum of mean squared errors for all resolutions.

It is worth highlighting that we do not assign different scale of persons to different levels in the feature pyramid, due to the following reasons. First, the heuristic used for assigning training target depends on both the dataset and network architecture. It is hard to transform the heuristic for FPN~\cite{lin2017feature} to HigherHRNet as both the dataset (scale distribution of person \emph{v.s.} all objects) and architecture (HigherHRNet only has 2 levels of pyramid while FPN has 4) change. Second, ground truth keypoint targets interact with each other since we apply the Gaussian kernel. Thus, it is very hard to decouple keypoints by simply setting ignored regions. We believe model has the ability to automatically focus on specific scales in different levels of the feature pyramid.

Tagmaps are trained differently from heatmaps in HigherHRNet. We only predict tagmaps at the lowest resolution, instead of using all resolutions. This is because learning tagmaps requires global reasoning and it is more suitable to predict tagmaps in lower resolution. Empirically, we also find higher resolutions do not learn to predict tagmaps well and even do not converge. Thus, we follow \cite{newwell2017associative} to train the tagmaps on feature maps at $\frac{1}{4}$ resolution of input image. 

\subsection{Heatmap Aggregation for Inference}
\label{sec:mr_inference}
We propose a heatmap aggregation strategy during inference. We use bilinear interpolation to upsample all the predicted heatmaps with different resolutions to the resolution of the input image and average the heatmaps from all scales for final prediction. This strategy is quite different from previous methods \cite{cao2017realtime,newwell2017associative,papandreou2018personlab} which only use heatmaps from a single scale or single stage for prediction.

The reason that we use heatmap aggregation is to enable scale-aware pose estimation. For example, the COCO Keypoint dataset \cite{lin2014microsoft} contains persons of large scale variance from $32^2$ pixels to more than $128^2$ pixels. Top-down methods~\cite{papandreou2017towards,chen2018cascaded,xiao2018simple} solve this problem by normalizing person regions approximately into a single scale. However, bottom-up methods need to be aware of scales to detect keypoints from all scales. We find heatmaps from different scales in HigherHRNet captures keypoints with different scales better. 
For example, keypoints for small persons missed in lower-resolution heatmap can be recovered in the higher-resolution heatmap.
Thus, averaging predicted heatmaps from different resolutions makes HigherHRNet a scale-aware pose estimator.

\begin{table*}[t]
\begin{threeparttable}
\centering
\setlength{\tabcolsep}{10.0pt}
\footnotesize
\begin{tabular}{c|c|c|c|c|ccccc}
Method & Backbone & Input size & \#Params & GFLOPs & 
$\operatorname{AP}$ & $\operatorname{AP}^{50}$ & $\operatorname{AP}^{75}$ & $\operatorname{AP}^{M}$ & $\operatorname{AP}^{L}$\\
\hline
\multicolumn{10}{c}{w/o multi-scale test}\\
\hline
OpenPose~\cite{cao2017realtime}\tnote{\textdagger} & - & - & - & - &$61.8$ & $84.9$ & $67.5$ & $57.1$ & $68.2$\\
Hourglass~\cite{newwell2017associative} & Hourglass & $512$ & $277.8$M& $206.9$ &$56.6$ & $81.8$&$61.8$&$49.8$&$67.0$\\
PersonLab~\cite{papandreou2018personlab} & ResNet-152 & $1401$ &$68.7$M& $405.5$ &$66.5$ & $88.0$&$72.6$&$62.4$&$72.3$\\
PifPaf~\cite{kreiss2019pifpaf} & - & - & - & - & $66.7$ & - & - & $62.4$ & $72.9$\\
Bottom-up HRNet\tnote{\textdaggerdbl}& HRNet-W32&$512$&$28.5$M&$38.9$
&$64.1$&$86.3$&$70.4$&$57.4$&$73.9$\\
HigherHRNet (Ours)& HRNet-W32&$512$&$28.6$M&$47.9$
&$66.4$&$87.5$&$72.8$&$61.2$&$74.2$\\
HigherHRNet (Ours)& HRNet-W48 &$640$&$63.8$M&$154.3$
&$\textbf{68.4}$&$\textbf{88.2}$&$\textbf{75.1}$&$\textbf{64.4}$&$\textbf{74.2}$\\
\hline
\multicolumn{10}{c}{w/ multi-scale test}\\
\hline
Hourglass~\cite{newwell2017associative} & Hourglass & $512$ & $277.8$M& $206.9$
&$63.0$ & $85.7$&$68.9$&$58.0$&$70.4$\\
Hourglass~\cite{newwell2017associative}\tnote{\textdagger} & Hourglass & $512$ & $277.8$M& $206.9$
&$65.5$ & $86.8$&$72.3$&$60.6$&$72.6$\\
PersonLab~\cite{papandreou2018personlab}  & ResNet-152 &$1401$ &$68.7$M& $405.5$
&$68.7$& $89.0$&$75.4$&$64.1$&$75.5$\\
HigherHRNet (Ours)& HRNet-W48&$640$&$63.8$M&$154.3$
&$\textbf{70.5}$&$\textbf{89.3}$&$\textbf{77.2}$&$\textbf{66.6}$&$\textbf{75.8}$\\
\end{tabular}
\begin{tablenotes}
\item[\textdagger] {\scriptsize Indicates using refinement.}
\item[\textdaggerdbl] {\scriptsize Our implementation, not reported in \cite{sun2019deep,WangSCJDZLMTWLX19}}
\end{tablenotes}
\end{threeparttable}
\caption{Comparisons with bottom-up methods on the \textbf{COCO2017 test-dev} set. All GFLOPs are calculated at single-scale. For PersonLab~\cite{papandreou2018personlab}, we only calculate its backbone's \#Params and GFLOPs. Top: w/o multi-scale test. Bottom: w/ multi-scale test. \emph{It is worth noting that our results are achieved without refinement.}}
\label{table:coco_test_dev_bu}
\end{table*}

\begin{table}[t]
    \centering
    \footnotesize
    \setlength{\tabcolsep}{2.0pt}
    \begin{tabular}{c|c|c|c|c|c|c}
    Method &  $\operatorname{AP}$ & $\operatorname{AP}^{50}$ & $\operatorname{AP}^{75}$ & $\operatorname{AP}^{M}$ & $\operatorname{AP}^{L}$ & $\operatorname{AR}$\\
    \hline
    \multicolumn{7}{c}{Top-down methods}\\
    \hline
    Mask-RCNN~\cite{he2017mask}& $63.1$ & $87.3$&$68.7$&$57.8$&$71.4$&-\\
    G-RMI~\cite{papandreou2017towards} &$64.9$ & $85.5$&$71.3$&$62.3$&$70.0$&$69.7$\\
    Integral Pose Regression~\cite{sun2018integral}  &$67.8$ & $88.2$&$74.8$&$63.9$&$74.0$&-\\
    G-RMI + extra data~\cite{papandreou2017towards} &$68.5$ & $87.1$&$75.5$&$65.8$&$73.3$&$73.3$\\
    CPN~\cite{chen2018cascaded}  & $72.1$ & $91.4$&$80.0$&$68.7$&$77.2$&$78.5$\\
    RMPE~\cite{fang2017rmpe}  &$72.3$ & $89.2$&$79.1$&$68.0$&$78.6$&-\\
    CFN~\cite{huang2017coarse}  & $72.6$ & $86.1$&$69.7$&$78.3$&$64.1$&-\\
    CPN (ensemble)~\cite{chen2018cascaded}  &$73.0$ & $91.7$&$80.9$&$69.5$&$78.1$&$ 79.0$\\
    SimpleBaseline~\cite{xiao2018simple} &${73.7}$ & ${91.9}$&${81.1}$&${70.3}$&${80.0}$&${79.0}$\\
    HRNet-W$48$~\cite{sun2019deep,WangSCJDZLMTWLX19} & ${75.5}$&${92.5}$&${83.3}$&${71.9}$&${81.5}$&${80.5}$\\
    HRNet-W$48$ + extra data~\cite{sun2019deep,WangSCJDZLMTWLX19} & ${77.0}$&${92.7}$&${84.5}$&${73.4}$&${83.1}$&${82.0}$\\ \hline
    \multicolumn{7}{c}{Bottom-up methods}\\
    \hline
    OpenPose$^{*}$~\cite{cao2017realtime} &$61.8$ & $84.9$&$67.5$&$57.1$&$68.2$&$66.5$\\
    Hourglass$^{*+}$~\cite{newwell2017associative}  &$65.5$ & $86.8$&$72.3$&$60.6$&$72.6$&$70.2$\\
    PifPaf~\cite{kreiss2019pifpaf} & 66.7 & - & - & 62.4 & 72.9 & - \\
    SPM~\cite{nie2019single} & 66.9 & 88.5 & 72.9 & 62.6 & 73.1 & - \\
    PersonLab$^{+}$~\cite{papandreou2018personlab}  &$68.7$ & $89.0$&$75.4$&$64.1$&$75.5$&$75.4$\\
    Ours: HigherHRNet-W48$^{+}$ &$70.5$&$89.3$&$77.2$&$66.6$&$75.8$&$74.9$\\
    \end{tabular}\vspace{2mm}
    \caption{Comparisons with both top-down and bottom-up methods on \textbf{COCO2017 test-dev} dataset. $^{*}$ means using refinement. $^{+}$ means using multi-scale test.} 
    \label{tab:coco_test_dev_all}
    \vspace{-2mm}
\end{table}

\section{Experiments}
\subsection{COCO Keypoint Detection}
\vspace{-2mm}

\paragraph{Dataset.} The COCO dataset \cite{lin2014microsoft} contains over $200,000$ images and $250,000$ person instances labeled with $17$ keypoints. COCO is divided into \textit{train}/\textit{val}/\textit{test-dev} sets with $57$k, $5$k and $20$k images respectively. All the experiments in this paper are trained only on the \textit{train} set. We report results on the \textit{val} set for ablation studies and compare with other state-of-the-art methods on the \textit{test-dev} set.
\vspace{-4mm}
\paragraph{Evaluation metric.}
The standard evaluation metric
is
based on Object Keypoint Similarity (OKS):
$\operatorname{OKS} = \frac{\sum_{i}\exp(-d_i^2/2s^2k_i^2)\delta(v_i > 0)}{\sum_i \delta(v_i > 0)}.$
Here $d_i$ is the Euclidean distance between 
a detected keypoint and its corresponding ground truth,
$v_i$ is the visibility flag of the ground truth,
$s$ is the object scale, and 
$k_i$ is a per-keypoint constant that controls falloff.
We report standard average precision and recall scores\footnote{\url{http://cocodataset.org/\#keypoints-eval}}:
$\operatorname{AP}^{50}$ ($\operatorname{AP}$ at $\operatorname{OKS} = 0.50$),
$\operatorname{AP}^{75}$,
$\operatorname{AP}$ 
(the mean of $\operatorname{AP}$ scores at 
$\operatorname{OKS} = 0.50, 0.55, \dots,0.90, 0.95$), 
$\operatorname{AP}^M$  for medium objects,
$\operatorname{AP}^L$  for large objects,
and $\operatorname{AR}$ (the mean of recalls at $\operatorname{OKS} = 0.50, 0.55, \dots,0.90, 0.95$). 

\begin{table}[bpt]
    \centering
    \setlength{\tabcolsep}{2.0pt}
    \begin{tabular}{c|c|ccc}
    Method&  Feat. stride/resolution& AP &$\text{AP}^{M}$&$\text{AP}^{L}$\\
    \hline
    HRNet&$4/128$&$64.4$&$57.1$&$75.6$\\ 
    HigherHRNet&$2/256$&$66.9$&$61.0$&$75.7$\\ 
    HigherHRNet&$1/512$&$66.5$&$61.1$&$74.9$\\ 
    \end{tabular}\vspace{2mm}
    \caption{Ablation study of HRNet \vs HigherHRNet on \textbf{COCO2017 val} dataset. Using one deconvolution module for HigherHRNet performs best on the COCO dataset.}
    \label{tab:ablation_highernet}
    \vspace{-2mm}
\end{table}
\vspace{-4mm}
\paragraph{Training.} Following \cite{newwell2017associative}, we use data augmentation with random rotation ([$-\ang{30}$, $\ang{30}$]), random scale ([$0.75$, $1.5$]), random translation ([$-40$, $40$]) to crop an input image patch of size $512\times512$ as well as random flip. As mentioned in Section~\ref{sec:mr_supervision}, we generate two ground truth heatmaps with resolutions $128\times128$ and $256\times256$ respectively.

We use the Adam optimizer \cite{kingma2014adam}. The base learning rate is set to $1e-3$, and dropped to $1e-4$ and $1e-5$ at the $200th$ and $260th$ epochs respectively. We train the model for a total of $300$ epochs. To balance the heatmap loss and the grouping loss, we set the weight to $1$ and $1e-3$ respectively for the two losses.
\paragraph{Testing.} We first resize the short side of the input image to 512 and keep the aspect ratio. Heatmap aggregation is done by resizing all the predicted heatmaps to the size of input image and taking the average. Following~\cite{newwell2017associative}, flip testing is used for all the experiments. All reported numbers have been obtained with single model without ensembling.
\vspace{-2mm}
\paragraph{Results on COCO2017 test-dev.} Table~\ref{table:coco_test_dev_bu} summarizes the results on COCO2017 test-dev dataset. From the results, we can see that using HRNet~\cite{sun2019deep,WangSCJDZLMTWLX19} itself already serves as a simple and strong baseline for bottom-up methods ($64.1$ AP). Our baseline method of HRNet with only single scale test outperforms Hourglass~\cite{newwell2017associative} using multi-scale test, while HRNet has much less parameters and computation in terms of FLOPs. Equipped with light-weight deconvolution modules, our proposed HigherHRNet ($66.4$ AP) outperforms HRNet by $+2.3$ AP with only marginal increase in parameters (+$0.4\%$) and FLOPs (+$23.1\%$). HigherHRNet is comparable with PersonLab \cite{papandreou2018personlab} but with only $50\%$ parameters and $11\%$ FLOPs. If we further use multi-scale test, our HigherHRNet achieves $70.5$ AP, outperforming all existing bottom-up methods by a large margin. We do not use any post processing like refining with top-down methods in~\cite{cao2017realtime,newwell2017associative}.

Table~\ref{tab:coco_test_dev_all} lists both bottom-up and top-down methods on the COCO2017 test-dev dataset. HigherHRNet further closes the performance gap between bottom-up and top-down methods.

\subsection{Ablation Experiments}
\begin{figure*}[ht]
    \centering
    \includegraphics[width=0.98\textwidth]{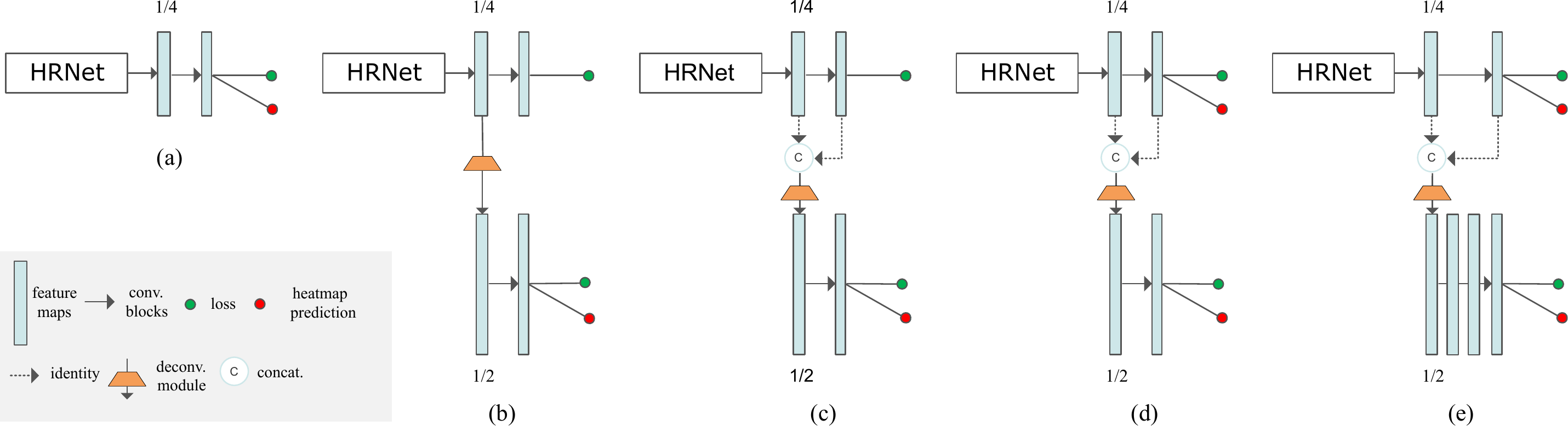}
    \caption{(a) Baseline method using HRNet~\cite{sun2019deep,WangSCJDZLMTWLX19} as backbone. (b) HigherHRNet with multi-resolution supervision~(MRS). (c) HigherHRNet with MRS and feature concatenation. (d) HigherHRNet with MRS and feature concatenation. (e) HigherHRNet with MRS, feature concatennation and extra residual blocks. For (d) and (e), heatmap aggregation is used.}
    \label{fig:ablation}
\end{figure*}
\begin{table*}[bpt]
    \centering
    \begin{tabular}{c|c|c|c|c|c|c|c|c}
    &Network& w/ MRS &feature concat. & w/ heatmap aggregation & extra res. blocks  &AP &$\text{AP}^{M}$&$\text{AP}^{L}$\\
    \hline
    (a)& HRNet& & && &$64.4$ &$57.1$&$75.6$\\
    (b)& HigherHRNet& \checkmark& &&&$66.0$ &$60.7$&$74.2$\\
    (c)& HigherHRNet& \checkmark& \checkmark&&&$66.3$ &$60.8$&$74.0$\\
    (d)& HigherHRNet& \checkmark&\checkmark &\checkmark &&$66.9$&$61.0$&$75.7$\\
    (e)& HigherHRNet& \checkmark&\checkmark &\checkmark &\checkmark&$67.1$&$61.5$&$76.1$\\
    \end{tabular}\vspace{2mm}
    \caption{Ablation study of HigherHRNet's components on \textbf{COCO2017 val} dataset. MSR: multi-resolution supervision. feature concat.: feature concatenation. res. blocks: residual blocks. }
    \label{tab:ablation_multi_supervision}
\end{table*}
We perform a number of ablation experiments to analyze Scale-Aware High-Resolution Network (HigherHRNet) on the COCO2017~\cite{lin2014microsoft} val dataset.
\vspace{-4mm}
\paragraph{HRNet \vs HigherHRNet.}

We perform ablation study comparing HRNet and HigherHRNet. For HigherHRNet, deconvolution module without extra residual blocks is used, and heatmaps aggregation is used for inference. Results are shown in Table~\ref{tab:ablation_highernet}. A simple bottom-up baseline by using HRNet with a feature stride of $4$ achieves $\text{AP}=64.4$. By adding one deconvolution module, our HigherHRNet with a feature stride of $2$ outperforms HRNet by a large margin of $+2.5$ AP (achieving $66.9$ AP). Furthermore, the main improvement comes from medium persons, where $\text{AP}^{M}$ is improved from $57.1$ for HRNet to $61.0$ for HigherHRNet.

These results show that HigherHRNet performs much better with small scales thanks to its higher resolution heatmaps. We also find the AP for large person pose does no drop. This is mainly because we also use smaller resolution heatmaps for prediction. It demonstrates that 1) making prediction at higher resolution is beneficial to bottom-up pose estimation and 2) scale-aware prediction is important.

If we add a sequence of two deconvolution modules after HRNet to generate feature maps that is of the same resolution as the input image, we observe that the performance decreases to $66.5$ AP from $66.9$ AP for adding only one deconvolution module. The improvement for medium person is marginal ($+0.1$ AP) but there is a large drop in the performance of large person ($-0.8$ AP). We hypothesize this is because the misalignment between feature map scale and object scales. Larger resolution feature maps (feature stride $=1$) are good for detecting keypoints from even smaller persons but the small persons in COCO are not considered for pose estimation. Therefore, we only use one deconvolution module by default for the COCO dataset. But we would like to point out that the number of cascaded deconvolution modules should be dependent on datasets and we will validate this on more datasets in our future work.
\vspace{-4mm}
\paragraph{HigherHRNet gain breakdown.}

To better understand the gain of the proposed components, we perform detailed ablation studies on each individual component. Figure~\ref{fig:ablation} illustrates all the architectures of our experiments. Results are shown in Table~\ref{tab:ablation_multi_supervision}.

\noindent\emph{Effect of deconvolution module.}~We perform ablation study on the effect of adding deconvolution module to generate higher resolution heatmaps. For a fair comparison, we only use the highest resolution feature maps to generate heatmaps for prediction (Figure~\ref{fig:ablation}~(b)). HRNet (Figure~\ref{fig:ablation}~(a)) achieves a baseline of $64.4$ AP. By adding one deconvolution module, the model achieves $66.0$ AP which is $1.6$ AP better than the baseline. This improvement is completely due to predicting on larger feature maps with higher quality. The result verifies our claim that it is important to predict on higher resolution feature maps for bottom-up pose estimation. 

\noindent\emph{Effect of feature concatenation.}~We concatenate feature maps with predicted heatmaps from HRNet as input to the deconvolution module (Figure~\ref{fig:ablation}~(c)) and the performance is further improved to $66.3$ AP. We also observe there is a large gain in medium persons while the performance for large persons decreases. Comparing method (a) and (c), the gain of predicting heatmaps at higher resolution mainly comes from medium persons ($+3.7 \text{AP}^{M}$). Moreover, the drop in large persons ($-1.6$ AP) justifies our claim that different different resolutions of feature maps are sensitive to different scales of persons.

\noindent\emph{Effect of heatmap aggregation.}~We further use all resolutions of heatmaps following the heatmap aggregation strategy for inference (Figure~\ref{fig:ablation}~(d)).  Compared with Figure~\ref{fig:ablation}~(c) ($66.3$ AP) that only uses the highest resolution heatmaps for inference, applying heatmap aggregation strategy achieves $66.9$ AP. Comparing method (d) and (e), the gain of heatmap aggregation comes from large person ($+1.7$ AP). And the performance of large person is even marginally better than predicting at lower resolution (method (a)). It means that predicting heatmaps using heatmap aggregation strategy is truly scale-aware.

\noindent\emph{Effect of extra residual blocks.}~We add 4 residual blocks in the deconvolution module and our best model achieves $67.1$~AP. Adding residual blocks can further refine the feature maps and it increases AP for both medium and large persons equally.

\vspace{-4mm}
\paragraph{Training with larger image size.}
A natural question is can training with larger input size further improve performance? To answer this question, we train HigherHRNet with $640\times640$ and $768\times768$ and results are shown in Table~\ref{tab:ablation_training_size}, all three models are tested using the training image size. We find that by increasing training image size to $640$, there is a significant gain of $1.4$ AP. Most of the gain comes from medium person while the performance of large person degrades slightly. When we further change the training image size to $768$, the overall AP does not change anymore. We observe a marginal improvement in medium person along with large degradation in large person. 

\begin{table}[bpt]
    \centering
    \setlength{\tabcolsep}{2.0pt}
    \begin{tabular}{c|ccc}
    Training size& AP &$\text{AP}^{M}$&$\text{AP}^{L}$\\
    \hline
    $512$&$67.1$&$61.5$&$76.1$\\ 
    $640$&$68.5$&$64.3$&$75.3$\\ 
    $768$&$68.5$&$64.9$&$73.8$\\ 
    \end{tabular}\vspace{2mm}
    \caption{Ablation study of HigherHRNet with different training image size on \textbf{COCO2017 val} dataset.}
    \label{tab:ablation_training_size}
\end{table}

\vspace{-4mm}
\paragraph{Larger backbone.}
In previous experiments, we use HRNet-W32 (1/4 resolution feature map has 32 channels) as backbone. We perform experiments with larger backbones HRNet-W40 and HRNet-W48. Results are shown in Table~\ref{tab:ablation_backbone}. We find using larger backbone consistently improves performance for both medium and large person.

\begin{table}[bpt]
    \centering
    \setlength{\tabcolsep}{2.0pt}
    \begin{tabular}{c|c|c|ccc}
    Backbone& \#Params & GFLOPs &AP& $\text{AP}^{M}$&$\text{AP}^{L}$\\
    \hline
    HRNet-W32&$28.6$&$47.8$&$68.5$&$64.3$&$75.3$\\ 
    HRNet-W40&$44.5$&$110.7$&$69.2$&$64.9$&$75.9$\\ 
    HRNet-W48&$63.8$&$154.3$&$69.9$&$65.4$&$76.4$\\ 
    \end{tabular}\vspace{2mm}
    \caption{Ablation study of HigherHRNet with different backbone on \textbf{COCO2017 val} dataset.}
    \label{tab:ablation_backbone}
    \vspace{-2mm}
\end{table}

\subsection{CrowdPose}
The CrowdPose~\cite{li2019crowdpose} dataset consists of 20,000 images, containing about 80,000 persons. The training, validation and testing subset are split in proportional to 5:1:4. CrowdPose has more crowded scenes than the COCO keypoint dataset, posing more challenges to pose estimation methods. The evaluation metric is the same as COCO~\cite{lin2014microsoft}.

The strong assumption of top-down methods that each person detection only contains a single person in the center, is no more valid in crowded scene. As shown in Table~\ref{tab:crowd_pose_test}, top-down methods~\cite{he2017mask,fang2017rmpe} that perform well on COCO fail on the CrowdPose dataset.

On the other hand, bottom-up methods naturally have the advantage in crowded scene. To validate the robustness of HigherHRNet in crowded scene, as well as setting up a strong baseline for bottom-up methods. We train our best HigherHRNet-W48 model on the CrowdPose \textit{train and val set} and report performance on the \textit{test} set. All training parameters follow COCO exactly and we use a crop size of $640\times640$ for both training and testing.

Results are shown in Table~\ref{tab:crowd_pose_test}. Our HigherHRNet outperforms na\"ive top-down methods by a large margin of 6.6 AP. HigherHRNet also outperforms the previous best method~\cite{li2019crowdpose} (which performs a global refinement of top-down method~\cite{fang2017rmpe}) by a healthy margin of 1.6 AP and most of the gain comes from $\operatorname{AP}^{M}$ (+1.8 AP) and $\operatorname{AP}^{H}$ (+1.5 AP), which contains images with the most crowd. Even without multi-scale test, HigherHRNet outperforms SPPE~\cite{li2019crowdpose} by 0.5 in $\operatorname{AP}^{H}$.

\begin{table}[t]
    \centering
    \footnotesize
    \setlength{\tabcolsep}{2.0pt}
    \begin{tabular}{c|c|c|c|c|c|c}
    Method &  $\operatorname{AP}$ & $\operatorname{AP}^{50}$ & $\operatorname{AP}^{75}$ & $\operatorname{AP}^{E}$ & $\operatorname{AP}^{M}$ & $\operatorname{AP}^{H}$\\
    \hline
    \multicolumn{7}{c}{Top-down methods}\\
    \hline
    Mask-RCNN~\cite{he2017mask}& $57.2$ & $83.5$ & $60.3$ & $69.4$ & $57.9$ & $45.8$ \\
    AlphaPose~\cite{fang2017rmpe} & $61.0$ & $81.3$ & $66.0$ & $71.2$ & $61.4$ & $51.1$ \\
    \hline
    \multicolumn{7}{c}{Top-down with refinement}\\
    \hline
    SPPE~\cite{li2019crowdpose} & $66.0$ & $84.2$ & $71.5$ & $75.5$ & $66.3$ & $57.4$ \\
    \hline
    \multicolumn{7}{c}{Bottom-up methods}\\
    \hline
    OpenPose~\cite{cao2017realtime} & - & - & - & $62.7$ & $48.7$ & $32.3$ \\
    Ours: HigherHRNet-W48 & $65.9$ & $86.4$ & $70.6$ & $73.3$ & $66.5$ & $57.9$ \\
    Ours: HigherHRNet-W48$^{+}$ & $67.6$ & $87.4$ & $72.6$ & $75.8$ & $68.1$ & $58.9$ \\
    \end{tabular}\vspace{2mm}
    \caption{Comparisons with both top-down and bottom-up methods on \textbf{CrowdPose test} dataset. Superscripts E, M, H of AP stand for easy, medium and hard. $^{+}$ means using multi-scale test.
    } 
    \label{tab:crowd_pose_test}
    \vspace{-2mm}
\end{table}

\vspace{-2mm}
\section{Conclusion}
We have presented a Scale-Aware High-Resolution Network (HigherHRNet) to solve the scale variation challenge in 
the bottom-up multi-person pose estimation problem, 
especially for precisely
localizing keypoints of small persons.
We find multi-scale image pyramid and larger input size partially solve the problem, but these methods suffer from high computational cost. 
To solve the problem, we propose an efficient high-resolution feature pyramid based on HRNet
and train it with multi-resolution supervision. 
During the inference, HigherHRNet with multi-resolution 
heatmap aggregation is capable of efficiently generating muilt- and higher-resolution heatmaps for more accurate human pose estimation.
HigherHRNet outperforms all existing bottom-up methods by a large margin on the challenging COCO dataset, especially for small persons.

{\small
\bibliographystyle{ieee_fullname}
\bibliography{egbib}
}

\end{document}